\documentclass[conference]{IEEEtran}
\IEEEoverridecommandlockouts
\usepackage{cite}
\usepackage{amsmath,amssymb,amsfonts}
\usepackage{algorithmic}
\usepackage{graphicx}
\usepackage{textcomp}
\usepackage{xcolor}
\usepackage[hyphens]{url}
\usepackage{hyperref}
\def\BibTeX{{\rm B\kern-.05em{\sc i\kern-.025em b}\kern-.08em
    T\kern-.1667em\lower.7ex\hbox{E}\kern-.125emX}}
\usepackage[flushleft]{threeparttable}
\begin{document}

\title{Domain Adaptation for sEMG-based Gesture Recognition with Recurrent Neural Networks}

\author{\IEEEauthorblockN{Istv\'an Ketyk\'o, Ferenc Kov\'acs and Kriszti\'an Zsolt Varga}
\IEEEauthorblockA{\textit{Nokia Bell Labs} \\
\{istvan.ketyko, ferenc.2.kovacs, krisztian.varga\}@nokia-bell-labs.com}
}

\maketitle

\IEEEpubid{\begin{minipage}{\textwidth}\ \\[12pt] \centering
\copyright 2019 IEEE. Personal use of this material is permitted. Permission from IEEE must be obtained for all other uses, in any current or future
media, including reprinting/republishing this material for advertising or promotional purposes, creating new collective works, for resale or redistribution to servers or lists, or reuse of any copyrighted component of this work in other works. DOI: \href{https://doi.org/10.1109/IJCNN.2019.8852018}{10.1109/IJCNN.2019.8852018}.
\end{minipage}}

\begin{abstract}
Surface Electromyography (sEMG/EMG) is to record muscles' electrical activity from a restricted area of the skin by using electrodes. The sEMG-based gesture recognition is extremely sensitive of inter-session and inter-subject variances. We propose a model and a deep-learning-based domain adaptation method to approximate the domain shift for recognition accuracy enhancement. Analysis performed on sparse and HighDensity (HD) sEMG public datasets validate that our approach outperforms state-of-the-art methods.
\end{abstract}


\section{Introduction} \label{sec:introduction}
\IEEEpubidadjcol
Traditionally, the control of a graphical user interface of a computer or the actions of a robot or drone is being done with hand or arm gestures interacting with a physical controller, like a mouse in case of traditional 2D screens. A touch sensor in case of touch screens can also be thought of as a physical controller. The wearable devices can keep the possibility to build a Human-Computer Interface (HCI), which gives an universal, natural and easy to use interaction with machines. With the advent of wearables there is a opportunity to get rid of the physical controller and interact with the computer without a proxy.

Sensing hand gestures without a physical proxy can be done by means of wearables or by means of image or video analysis of hand or finger motion \cite{b1}. A wearable-based detection can physically rely on measuring the acceleration and rotations of our bodily parts (arms, hands or fingers) with Inertial Measurement Unit (IMU) sensor(s) or by measuring the myo-electric signals generated by the various muscles of our arms or fingers with EMG sensors. Surface EMG (sEMG) records muscle activity from the surface of the skin which is above the muscle being evaluated. The signal is collected via surface electrodes. Both type of sensors have their own application areas:
\begin{itemize}
	\item IMU sensors are typically used for detecting large movements and they are not suitable for recognizing fine gestures such as spread fingers or finger pinching,
	\item EMG devices are typically used for gesture recognition.
\end{itemize}
Usually, both types of sensor data are needed to have good user experience from a HCI point of view. 

In this paper we focus on the main challenges of sEMG-based gesture recognition. In fact, this translates to a time series classification task and several papers provide solutions from classic data science solutions \cite{b7} to deep learning classifications \cite{b18} and it is an active research topic. sEMG signals highly depend on:
\begin{itemize}
 \item The subject under test,
 \item Physical conditions of the subject (e.g., skin conductivity),
 \item External/measurement conditions (e.g., sensor placement accuracy).
 \end{itemize} 
 
If these dependencies are not taken into consideration, like in a scenario when gestures are recognized for one subject in one session without the device being removed from the surface of the skin, the accuracy of state of the art classifiers is above 90\%. If any or all these conditions are not met, the accuracy of  gesture recognition accuracy degrades to below 50\%. In this paper we propose a domain adaptation model that can handle them efficiently. 

This paper is organized as follows, Section~\ref{sec:related_work} provides a short summary of the used technologies,then our adaptation model introduced in Section~\ref{sec:architecture}. Next, we validate our approach using publicly available sEMG data sets: the experimental setup is described in Section~\ref{sec:experimental_setup}, and Section~\ref{sec:results} gives the detailed analysis of the experimentation. Finally, we conclude and summarize our results.
\IEEEpubidadjcol

\section{Related Work} \label{sec:related_work}

In this section, the used techniques and technologies are introduced:
\begin{itemize}
	\item Main properties of sEMG signals and sensors,
	\item sEMG-based gesture recognition techniques,
	\item Recurrent Neural Networks (RNN).
\end{itemize} 

\subsection{Gestures and sEMG}

The formation of a hand gesture usually adheres to the following pattern. In the onset stage the hand and/or fingers start to execute a motion from a relaxed position until the point they reach their final state because of a physical constraint, and then in the gesture termination stage they go back to a relaxed position. In a sequence of gestures, like in the case of sign languages, the relaxed position is not reached for a long time. Therefore most hand gestures may be considered mixed from the muscle contraction perspective. 

From the perspective of contraction pattern, hand gestures can determine muscles to be contracted in an isotonic, isometric or mixed pattern. Isotonic contractions involve muscular contractions against resistance in which the length of the muscle changes. Contrary to isotonic contractions, isometric contractions create no change in muscle length but tension and energy are fluctuating. An isometric contraction is typically performed against an immovable object \cite{b17}.

Furthermore, a sequence of gestures is always a sequence of one or two isotonic contractions followed by exactly one isometric contraction. This is because of the following: during the time interval between relaxed and final states the contraction type is isotonic, because the length of the muscle changes. During the time period where the final state of the gesture is maintained, the contraction type is isometric: tension and energy may fluctuate, but the length of the muscle stays stationary. Finally, in the period when the gesture is terminated and a new gesture follows in the sequence the contraction type is isotonic: the hand/finger is released and/or immediately afterwards contracted again to form the new gesture.

Muscles generate electric voltage during contraction/detraction. EMG detectors measure this signal through electrodes that are attached to the skin. A digital-analogue conversion is performed with a sampling rate of 100 up to 2000 Hz and the outcome is usually normalized into a range of [-1.0, 1.0]. The typical bandwidth of this signal is 5-450 Hz \cite{emgbw}. This set of time series (one per each pair of electrodes) represents usually the input for gesture detection algorithms.

In the number of sensors point of view, two different types of measurement configuration are in use:
\begin{itemize}
	\item \textit{sparse EMG:} Only a couple of sensors are attached to the skin. Typically, 8- 10 sensors are used in this configuration. 
	\item \textit{dense EMG:} Tens of sensors are attached to the skin. Usually, these sensors are arranged in a matrix and they cover an area of the skin. If the number of sensors is more than 100, the configuration is called high density EMG.
\end{itemize}
Both configurations have pros and cons. On the one hand, sparse EMG need smaller bandwidth as it has fewer channels and less data to transfer, on the other hand, it is more sensitive to the sensor placement. Otherwise, the dense EMG setups are less sensitive to the sensor placement, but they need more bandwidth and it has wiring issues in case of wearable devices. There are several publicly available EMG datasets, some of them were recorded with parse sensor configuration, while others using dense setup. \cite{b6}, \cite{b7}, \cite{b8} and \cite{b12}.

\subsection{sEMG-based gesture detection}

\begin{figure}[tbp]
	\centerline{\includegraphics[width=0.8\columnwidth,keepaspectratio]{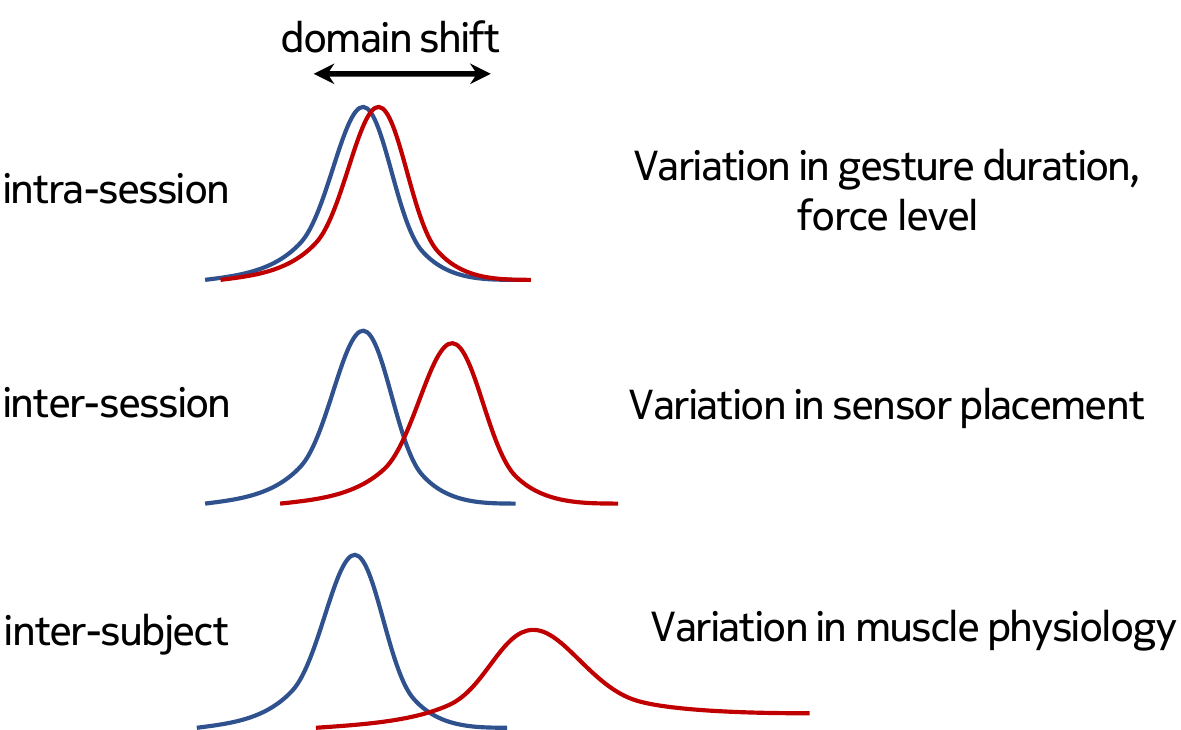}}
	\caption{Domain shift in case of different scenarios}
	\label{figure:domainShift}
\end{figure}

During a (recording/testing) session, there are several repetitions/trials of the same gesture set by the human subjects.
Meanwhile, the sEMG electrode sensors expected to remain in the same placement.
Multiple sessions naturally differ by sensor placement because of their shift on the skin and rotation around the arm.
Apparently, the sensor placement and electrode skin contact on different subjects has the highest alternation.
In sEMG-based gesture recognition there are three cases in terms of the data variability (as shown on Fig.~\ref{figure:domainShift}):
\begin{enumerate}
	\item Intra-session: in this case the data variabilty comes from differences between the trials/repetitions of the performed gestures by the human subject.
	\item Inter-session: in this case there is still the intra-session variability with an additional data variabilty which comes from the differences between the recording sessions.
	At each recording session the sensor placement can have some shift and/or rotations.
	\item Intra-subject: The electromyogram signal is a kind of biological signal which is severely affected by the
	difference between subjects. In this case the data variabilty comes from the differences of human subjects.
\end{enumerate}

The intra-session gesture recognition have been extensively researched.
Existing sEMG-based solutions utilizes time domain, frequency domain, and time-frequency domain features.
Many researchers focused on presenting new sEMG features based on their domain knowledge or analyzing existing features to propose new feature sets.
Traditional machine learning classifiers have been employed to recognize sEMG-based gestures, such as k-Nearest Neighbor (kNN) \cite{b20},
Linear Discriminate Analysis (LDA) \cite{b21}, Hidden Markov Model (HMM) \cite{b22}, and Support Vector Machine (SVM) \cite{b20}\cite{b23}.
The Convolutional Neural Network (CNN) architecture is the most widely used deep learning technique for sEMG-based gesture recognition.
\cite{b18} provided a novel CNN model to extract spatial information from the instantaneous sEMG images and achieved state-of-the-art performance.
\cite{b19} applies a novel hybrid CNN-RNN architecture with superior results in the intra-session scenario.

The inter-session and inter-subject variability causes domain shift in the distributions of the sEMG sensor data.
From a machine learning viewpoint, one of the key issues in inter-session/subject Muscle-Computer Interfaces (MCIs) is domain adaptation,
i.e., developing learning algorithms in which the training data (source domain) used to learn a model have a different distribution compared with
the data (target domain) to which the model is applied \cite{b18}.
Domain adaptation has gained increasing interest in the context of deep learning.
When only a small amount of labeled data is available in the target domain during the training phase,
fine-tuning pre-trained networks has become the de facto method.

Another approach is the unsupervised adaptive learning which utilises only unlabellet target data.
\cite{b20} compares four concepts which work with SVM and provides state-of-the-art results on the NinaPro dataset.
\cite{b18} provides the state-of-the-art solution on the CapgMyo dataset. They invented a multi-source adaptive batch normalization technique which works with CNN architecture.
The drawback of this solution, that in case of multiple sources (i.e., multiple subjects), constraints and considerations are needed per source at pre-training time of that model.

\subsection{Recurrent Neural Networks}

\begin{figure}[tbp]
\centerline{\includegraphics[width=200px,keepaspectratio]{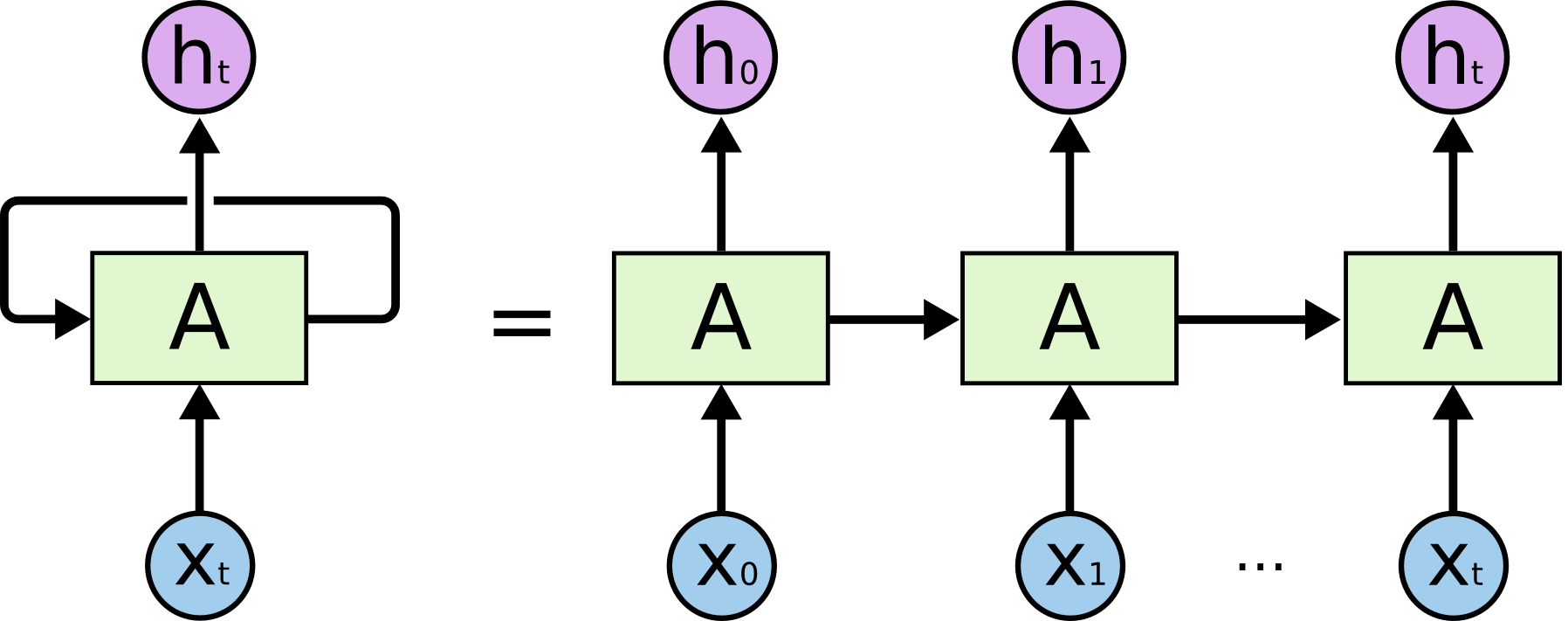}}
\caption{A simple RNN in compact and unrolled representation\cite{colahweb}}
\label{fig_rnn}
\end{figure}

With the increase of computational capabilities in the recent years, neural networks have become more popular due to their ability to tackle complex data science problems. A typical neural network has an input layer, one or many hidden layers and an output layer. Each hidden layer has a set of nodes that take in weighted inputs from the previous layer and provide an output through an activation function to the next layer. Recurrent neural networks (RNNs) are a family of neural networks in which there are feedback loops in the system. Feedback loops allow processing the previous output with the current input, thus making the network stateful, being influenced by or “remembering” the earlier inputs in each step (see Fig. \ref{fig_rnn}). A hidden layer that has feedback loops is also called a recurrent layer. The mathematical representation of a simple recurrent layer can be seen in Eq. (\ref{eq:rnn}).

\begin{equation} \label{eq:rnn}
\begin{aligned}
& \mathbf{h}_{t} = \sigma_{h} (\mathbf{w}_{h}\mathbf{x}_{t}+\mathbf{u}_{h}\mathbf{h}_{t-1}+\mathbf{b}_{n}) \\
& \mathbf{y}_{t} = \sigma_{y} (\mathbf{w}_{y}\mathbf{h}_{t}+\mathbf{b}_{y})
\end{aligned}
\end{equation}

However, regular RNNs suffer from the vanishing gradient problem which means that the gradient of the loss function decays exponentially with time, making it difficult to learn long-term temporal dependencies in the input data. \cite{razvan} Long Short Term Memory (LSTM) networks had been proposed to solve this problem. They are a special type of RNN that attempt to solve the vanishing gradient problem \cite{hochreiter}. In this paper we will present a solution that is utilizing LSTM cells. 

\begin{figure}[tbp]
\centerline{\includegraphics[width=200px,keepaspectratio]{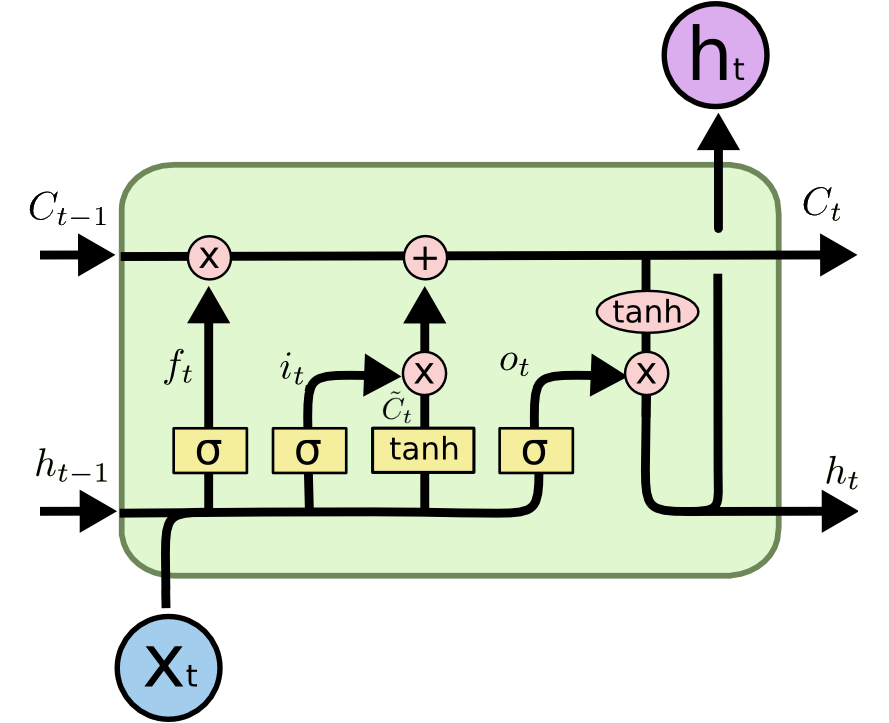}}
\caption{LSTM cell architecture\cite{colahweb}}
\label{lstm_arch}
\end{figure}

LSTM units contain a set of gates that are used to control the stages when information enters the memory (input gate: $\mathbf{i}_{t}$), when it's output (output gate: $\mathbf{o}_{t}$) and when it's forgotten (forget gate: $\mathbf{f}_{t}$) as seen in Eq. (\ref{eq:lstm}). This architecture allows the neural network to learn longer-term dependencies and they are widely used to analyze time-series data. \cite{chung} In Fig.~\ref{lstm_arch} yellow rectangles represent a neural network layer, circles are point-wise operations and arrows denote the flow of data. 

\begin{equation} \label{eq:lstm}
\begin{aligned}
& \mathbf{f}_{t} = \sigma( \mathbf{W}_{f} \cdot [ \mathbf{h}_{t-1}, \mathbf{x}_{t} ] + \mathbf{b}_{f} )  \\
& \mathbf{i}_{t} = \sigma( \mathbf{W}_{i} \cdot [ \mathbf{h}_{t-1}, \mathbf{x}_{t} ] + \mathbf{b}_{i} ) \\
& \mathbf{\widetilde{C}}_{t} = \tanh( \mathbf{W}_{C} \cdot [ \mathbf{h}_{t-1}, \mathbf{x}_{t} ] + \mathbf{b}_{C} ) \\
& \mathbf{C}_{t} = \mathbf{f}_{t} \ast \mathbf{C}_{t-1} + \mathbf{i}_{t} \ast \mathbf{\widetilde{C}}_{t} \\
& \mathbf{o}_{t} = \sigma( \mathbf{W}_{o} \cdot [ \mathbf{h}_{t-1}, \mathbf{x}_{t} ] + \mathbf{b}_{o} ) \\
& \mathbf{h}_{t} = \mathbf{o}_{t} \ast \tanh(\mathbf{C}_{t})
\end{aligned}
\end{equation}

\begin{figure}[tbp]
\centerline{\includegraphics[width=200px,keepaspectratio]{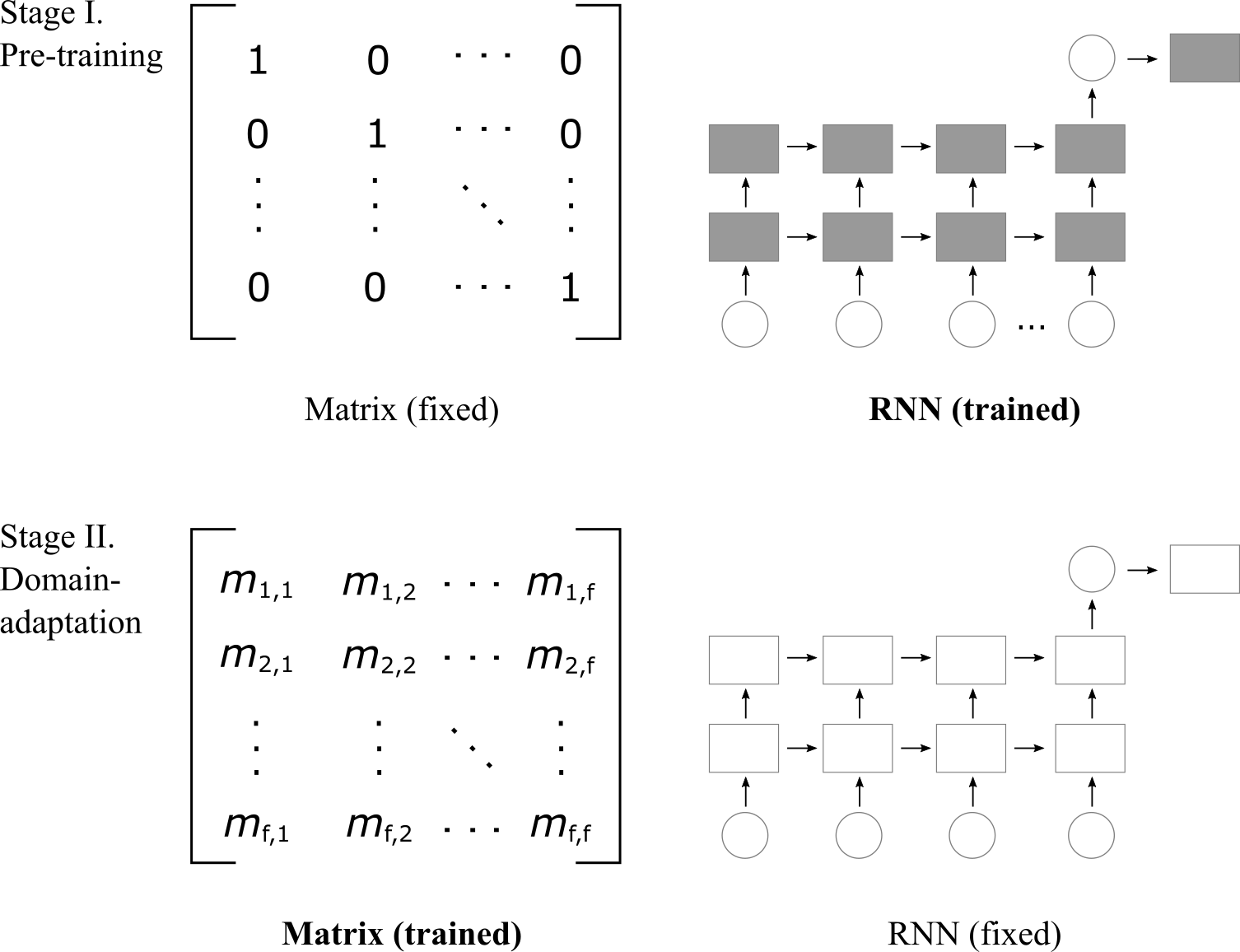}}
\caption{Our two-stage domain adaptation method}
\label{our_training}
\end{figure}

\begin{figure}[tbp]
\centerline{\includegraphics[width=200px,keepaspectratio]{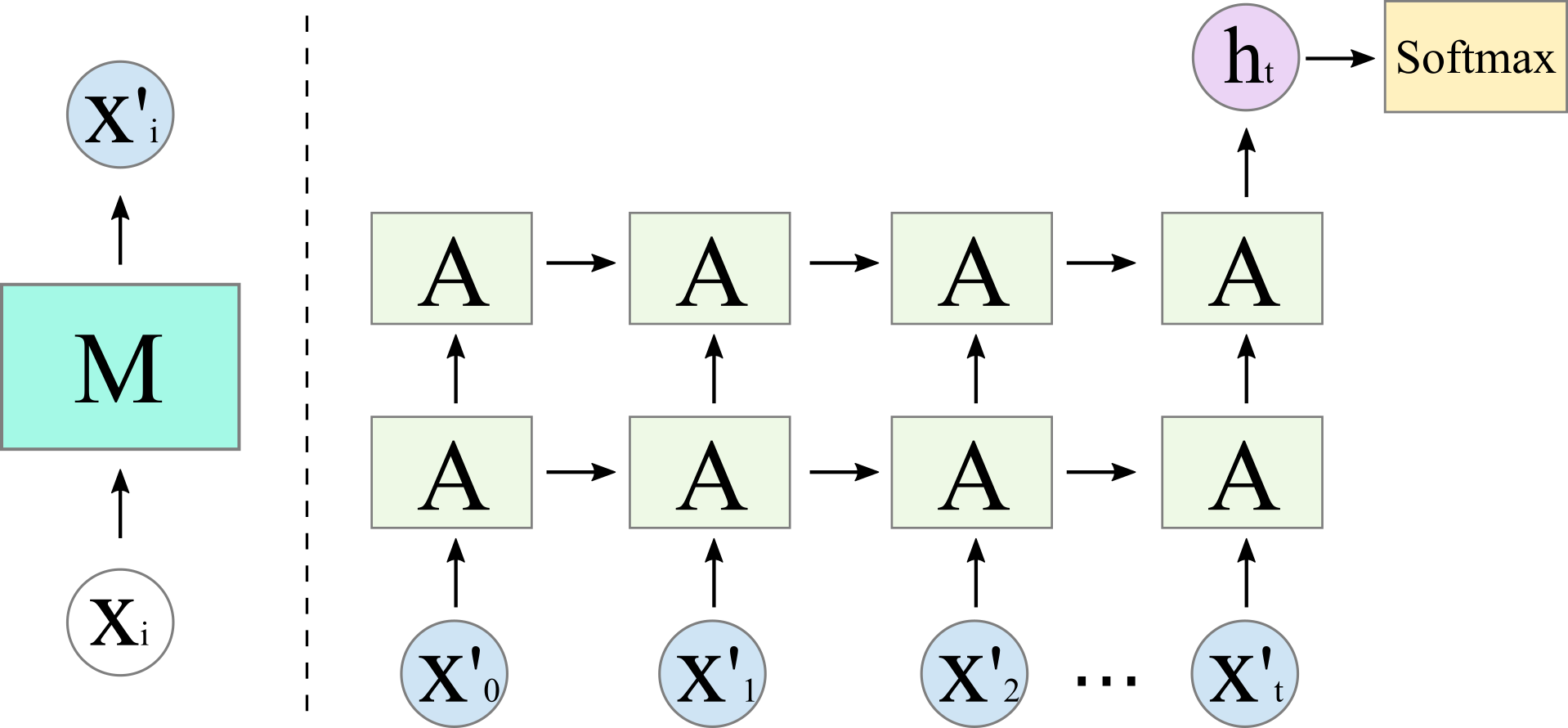}}
\caption{Our neural network architecture (2SRNN)}
\label{our_nw}
\end{figure}

\section{2-Stage Domain Adaptation} \label{sec:architecture}
\subsection{Architecture}

We propose a model which consists of two components as can be seen at Fig.~\ref{our_nw} and we name it as 2-Stage RNN (2SRNN):
\begin{enumerate}
  \item The domain adaptation layer: which is a single fully-connected layer without a non-linear activation function. The input vector $\mathbf{x} \in \mathcal{R}^f$ is the same dimension
  as the output vector $\mathbf{x'} \in \mathcal{R}^f$ where $f$ is the number of input features.
  The trainable weights form a square matrix $\mathbf{M} \in \mathcal{R}^{f \times f}$ plus there is a bias vector $\mathbf{b} \in \mathcal{R}^f$.

\def\horzbar{\text{magic}}
\begin{align*}
\begin{bmatrix}
  x'_{1}  \\
  \vdots \\
  x'_{f}
\end{bmatrix}
 &= \left[\begin{array}{ccc}
  m_{11} & \dots & m_{1f} \\
  \vdots & m_{ij} & \vdots \\
  m_{f1} & \dots & m_{ff} \\ 
\end{array}\right]
\begin{bmatrix}
  x_{1}  \\
  \vdots \\
  x_{f}
\end{bmatrix}
{} + \begin{bmatrix}
  b_{1}  \\
  \vdots \\
  b_{f}
\end{bmatrix}
\end{align*}

  \item The sequence classifier: which is a deep stacked RNN with many-to-one setup followed by a $G$-way fully-connected layer and a softmax classifier.
  $G$ is the number of gestures to be recognized.
\end{enumerate}

The linear transformation for domain adaptation with the same $\mathbf{M}$ and $\mathbf{b}$ is applied to the input of the RNN at each timestamp $t$: $\mathbf{x'_t}=\mathbf{M}\mathbf{x_t} + \mathbf{b}$.

The transformation of the input values (to solve the domain shift) is approximated with perceptron learning.
Our assumption is that this transformation is a linear one.
Apparently, a linear transformation yields the highest gain.
Also, it could be a more complex (polinomial or non-linear) one. There is still gain as long as the domain adaptation layer is smaller (in size and complexity) than the sequence classifier component.

\subsection{Method}

Fig.~\ref{our_training} visualises our method with two consecutive stages:

\subsubsection{Pre-training stage}
In the first stage, the weights of the domain adaptation layer are frozen and the sequence classifier is trained from scratch on the source dataset.
The domain adaptation layer's initial weights could be several combinations of real numbers but we chose $\mathbf{M}$ to be the identity matrix and $\mathbf{b}$ to be a vector of zeros to represent the identity transformation.
We apply supervised learning. The optimization is a gradient descent with backpropagation. The loss function is the categorical cross entropy:
\begin{equation}
  \label{eq:multipleentropy}
  \jmath_{\mbox{\tiny entropy}} = -\sum_{g \in G} {\textrm{I}_{g} \ln p_g}
\end{equation}
where $G$ is the number of gestures and I is the indicator function whether class label $g$ is the correct classification for the given observation and
$p$ is the predicted probability that the observation is of class $g$.

\subsubsection{Domain adaptation stage}
In the second stage, the weights of the sequence classifier are frozen in their pre-trained state and the domain adaptation layer's weights are trained on the target dataset.
In this stage, the same supervised learning is applied. The loss (Eq.~\eqref{eq:multipleentropy}) is backpropagated to the domain adaptation layer during the process.
The advantages of this architecture:
\begin{enumerate}
  \item The training during the second stage is very fast because there is only a shallow network to tackle with.
  \item Training a linear layer ensures the convergence.
\end{enumerate}

\section{Experimental setup} \label{sec:experimental_setup}

We approximate the inter-session and inter-subject shift in the values of the sEMG electrodes with a linear transformation of the input.
The transformation could be polinomial or non-linear also but that could result in either a larger domain adaptation network or a non-convex optimization manifold.
We let the model discover the coefficients of this linear transformation on the target data during the domain adaptation process.

We have tested the approach on HD sEMG and sparse sEMG datasets:
\begin{enumerate}
  \item CapgMyo dataset \cite{b18}: includes HD-sEMG data for 128 channels acquired from 23 intact subjects. The sampling rate is 1 KHz. It consists of 3 sub-databases:
  \begin{enumerate}
	\item DB-a: 8 isometric and isotonic hand gestures were obtained from 18 of the 23 subjects.
	\item DB-b: 8 isometric and isotonic hand gestures from 10 of the 23 subjects in two recording sessions on different days.
	\item DB-c: 12 basic movements of the fingers were obtained from 10 of the 23 subjects.
  \end{enumerate}
  We downloaded the pre-processed version from \url{http://zju-capg.org/myo/data} to use the same data as \cite{b18} to be able to compare our results with theirs.
  In that version, the power-line interference was removed from the sEMG signals by using a band-stop filter (45–55 Hz, second-order Butterworth).
  Only the static part of the movements was kept in it (for each trial, the middle one-second window, 1000 frames of data).
  They used the middle one second data to ensure that no transition movements are included in it.
  We rescaled the data to have zero mean and unit variance, then we rectified it and applied smoothing.
  \item NinaPro dataset \cite{b23}:
	  \begin{enumerate}
		\item DB1: The NinaPro sub-database 1 (DB-1) is for the development of hand prostheses, and contains sparse multi-channel sEMG recordings.
		It consists of a total of 52 gestures performed by 27 intact subjects.
		\begin{enumerate}
		\item Gesture numbers 1–12: 12 basic movements of the fingers (flexions and extensions). These are
		equivalent to gestures in CapgMyo DB-c.
		\item Gesture numbers 13–20: 8 isometric, isotonic hand configurations (”hand postures”). These are
		equivalent to gestures in CapgMyo DB-a and DB-b.
	  \end{enumerate}
  \end{enumerate}
  The data is recorded at a sampling rate of 100 Hz, using 10 sparsely located electrodes placed on subjects’ upper forearms.
  The sEMG signals were rectified and smoothed by the acquisition device.
  We downloaded the re-organized version from \url{http://zju-capg.org/myo/data/ninapro-db1.zip} to use the same data as \cite{b18} for fair comparison.
  For each trial, we used the middle 1.5-second window, 180 frames of data to get the static part of the movements.
  We used the middle 1.5-second data with the aim that no transition movements are included in it.
\end{enumerate}

We decompose the sEMG signals into small sequences using the sliding window strategy with overlapped windowing scheme.
The sequence length must be shorter than 300ms \cite{b19} to satisfy real-time usage constraints. To compare our proposed method with previous works, we follow
the segmentation strategy in previous studies.

We use Keras with Tensorflow backend. The domain adaptation layer has the $\mathbf{M} \in \mathcal{R}^{f \times f}$ where $f$ is 128 in case of the CapgMyo dataset and 10 in case of the NinaPro.
It is implemented with the TimeDistributed Keras wrapper.
For the sequence classifier we use a 2-stack RNN with LSTM cells. Each LSTM cell has a dropout with the probability of 0.5 and 512
hidden units. The RNN is followed by a $G$-way fully-connected layer with 512 units (dropout with a probability of 0.5) and a softmax classifier.
We use the Adam optimizer \cite{adam} with the learning rate of 0.001.

\section{Results} \label{sec:results}

The outcomes of our investigations are compared with results from \cite{b23}, \cite{b18}, \cite{b7}, and \cite{b19}.

Based on the classification run on the test dataset, taken from the same database as the training dataset in a manner detailed in the subsequent, the classification accuracy is calculated for each database as given below:
\begin{equation}
  \label{eq:classification_accuracy}
  \mbox{Classification Accuracy} = \dfrac{\mbox{Correct classifications}}{\mbox{Total classifications}}*100\%
\end{equation}

\subsection{Intra-session validation}

We first look at intra-session validation to benchmark our sequence classifier against the state-of-the-art in the least challenging scenario, on top of distinct datasets, without performing any further optimization.

In case of CapgMyo dataset we used the same evaluation procedure that was used in the previous study \cite{b18}.
For each subject, a classifier was trained by using 50\% of the data (E.g., trials 1, 3, 5, 7 and 9 for that subject) and tested by using the remaining half.
This procedure was performed on each sub-database. For DB-b, the second session of each subject was used for the evaluation.

In previous works on NinaPro DB-1 \cite{b23}, \cite{b18}, the training set consisted of approximately 2/3 of the gesture trials of each subject and the remaining trials constitute the test set.

We chose 150-ms sequence length for the RNN in all the cases for fair comparison. Table~\ref{table:intra-session} shows our average intra-session recognition accuracy together with the state-of-the-art.
Columns noted with DB-a, DB-b, DB-c belong to the CapgMyo dataset and columns noted with DB-1 12 gestures and DB-1 8 gestures belong to the NinaPro dataset.
\begin{table}[tbp]
	\centering
	\begin{threeparttable}
		\setlength{\tabcolsep}{3pt}
		\def\arraystretch{1.5}%
		\begin{tabular}{|c|c|c|c|c|c|} 
			\hline
			& \multicolumn{3}{|c|}{CapgMyo} & \multicolumn{2}{|c|}{NinaPro} \\
			\hline
			& DB-a & DB-b & DB-c & DB-1 12 gestures & DB-1 8 gestures \\
			Du\cite{b18} & 99.5\% & 98.6\% & 99.2\% & 84\% & 83\% \\
			Hu\cite{b19} & 99.7\% & -\tnote{a} & - & - & - \\
			Atzori\cite{b23} & - & - & - & 90\% & - \\
			2SRNN & 97.1\% & 97.1\% & 96.8\% & 84.7\% & 90.7\% \\
			\hline
		\end{tabular}
		\caption{Intra-session recognition accuracy results}
		\begin{tablenotes}
			\item[a] '-' notes that the authors of that method did not focus on the scenario.
		\end{tablenotes}
		\label{table:intra-session}
	\end{threeparttable}
\end{table}
As can be seen from Table~\ref{table:intra-session} the accuracy achieved by our model is at most 2.4 percentage points worse than other methods for the CapgMyo database and with 0.7 up to 7.7 percentage points better for the NinaPro dataset. This accuracy has been achieved by keeping the training duration to constant 100 epochs and without any hyper-parameter tuning. This outcome indicates that that our model is at least comparable in the intra-session case with other approaches.

\subsection{Inter-session validation}

We evaluated inter-session recognition for CapgMyo DB-b,
in which the model was trained using data recorded from the first session and evaluated using data recorded from the second session.
In each case, we ran our domain adaptation for 100 epochs using the following 3 scenarios:
\begin{enumerate}
	\item Scenario 1: domain adaptation is not applied,
	\item Scenario 2: domain adaptation performed on the complete set of target data (all data of the target session); this scenario has only been considered for the purpose of comparability with alternative approaches,
	\item Scenario 3: domain adaptation performed on 50\% of the trials of the target session, while the validation set is the remaining 50\%.
\end{enumerate}
Our adaptation scheme enhanced inter-session recognition with 29 percentage points (accuracy of 83.8\% compared to 54.6\%) which is a 53\% improvement (shown in Table~\ref{table:inter-session}). 

\begin{table}[tpb]
	\centering
	\begin{threeparttable}
		\def\arraystretch{1.5}%
		\begin{tabular}{|c|c|c|c|} 
			\hline
			& Scenario 1 & Scenario 2 & Scenario 3 \\
			\hline
			Du\cite{b18} & 47.9\% & - & 63.3\% \\
			2SRNN & 54.6\% & 85.8\% & 83.8\% \\
			\hline
		\end{tabular}
		\caption{Inter-session recognition accuracy results on CapgMyo DB-b}
		\label{table:inter-session}
	\end{threeparttable}
\end{table}

\subsection{Inter-subject validation}

In this experiment,
we evaluated inter-subject recognition of 8 gestures using the second recording session of CapgMyo DB-b and the recognition of 12 gestures using CapgMyo DB-c and the sub-set of 12 gestures from the NinaPro DB-1.
We performed a leave-one-out cross-validation,
in which each of the subjects was used in turn as the test subject and a classifier was trained using the data of the remaining subjects, using the following 3 scenarios:
\begin{enumerate}
	\item Scenario 1: domain adaptation is not applied,
	\item Scenario 2: domain adaptation performed on the complete set of target data (all data of the target subject); this scenario has only been considered for the purpose of comparability with alternative approaches,
	\item Scenario 3: domain adaptation performed on 50\% of the trials of the target subject, while the validation set is the remaining 50\%, with the following 2 variants:
	\begin{enumerate}
		\item CapgMyo DB-b, DB-c: 50\%-50\% of the target subject data (5 of the 10 trials are used for domain adaptation and another 5 is for its validation).
		\item NinaPro DB-1 12 gestures: 50\%-50\% of the target subject data (5 of the 10 trials are used for domain adaptation and another 5 is for its validation).
	\end{enumerate}
\end{enumerate}
In case of the CapgMyo DB-b and DB-c we ran our domain adaptation for 100 epochs, and in case of the Ninapro DB-1 for 400 epochs.
The sequence length of our RNN was 150 ms in case of the CapgMyo DB-b and DB-c for comparison reasons with \cite{b18}, and 400 ms in case of the Ninapro DB-1 for comparison reasons with \cite{b7}.
Table~\ref{table:inter-subject} shows the classification accuracies of the various methods.
\begin{table}[tpb]
	\centering
	\begin{threeparttable}
		\setlength{\tabcolsep}{1pt}
		\def\arraystretch{1.5}%
		\begin{tabular}{|c|c|c|c|c|c|c|c|c|c|} 
			\hline
			& \multicolumn{3}{|c|}{Scenario 1} & \multicolumn{3}{|c|}{Scenario 2} & \multicolumn{3}{|c|}{Scenario 3} \\
			\hline
			& DB-b & DB-c & DB-1 & DB-b & DB-c & DB-1 & DB-b & DB-c & DB-1 \\
			Du\cite{b18} & 39.0\% & 26.3\% & - & - & - & - & 55.3\% & 35.1\% & - \\
			Atzori\cite{b23} & - & - & 25\% & - & - & - & - & - & -\\
			Patricia\cite{b7} & - & - & 30\% & - & - & - & - & - & 55\% \\
			2SRNN & 52.6\% & 34.8\% & 35.1\% & 96.8\% & 91.9\% & 65.7\% & 89.9\% & 85.4\% & 65.2\% \\
			\hline
		\end{tabular}
		\caption{Inter-subject recognition accuracy results}
		\label{table:inter-subject}
	\end{threeparttable}
\end{table}
Our adaptation scheme enhanced inter-subject recognition with a 71\% improvement on DB-b, 145\% improvement on DB-c and 86\% improvement on DB-1 12 gestures (shown in Table~\ref{table:inter-subject}).

We summarise the domain adaptation improvement results in Table~\ref{table:speed-up}. As indicated there, the performance of 2SRNN is superior in all cases: the improvement obtained from our domain adaptation in the inter-session and inter-subject cases exceeds those obtained through alternative domain adaptation approaches.
\begin{table}[tpb]
	\centering
	\begin{threeparttable}
		\setlength{\tabcolsep}{2pt}
		\def\arraystretch{1.5}%
		\begin{tabular}{|c|c|c|c|c|c|} 
			\hline
			& Inter-session improvement & \multicolumn{3}{|c|}{Inter-subject improvement} \\
			\hline
			& DB-b & DB-b & DB-c & DB-1 \\
			Du\cite{b18} & 32\% & 42\% & 33\% & - \\
			Patricia\cite{b7} & - & - & - & 83\% \\
			2SRNN & \textbf{53\%} & \textbf{71\%} & \textbf{145\%} & \textbf{86\%} \\
			\hline
		\end{tabular}
		\caption{Domain adaptation improvement comparisons}
		\label{table:speed-up}
	\end{threeparttable}
\end{table}

\begin{figure}[tbp]
	\centerline{\includegraphics[width=\columnwidth,keepaspectratio]{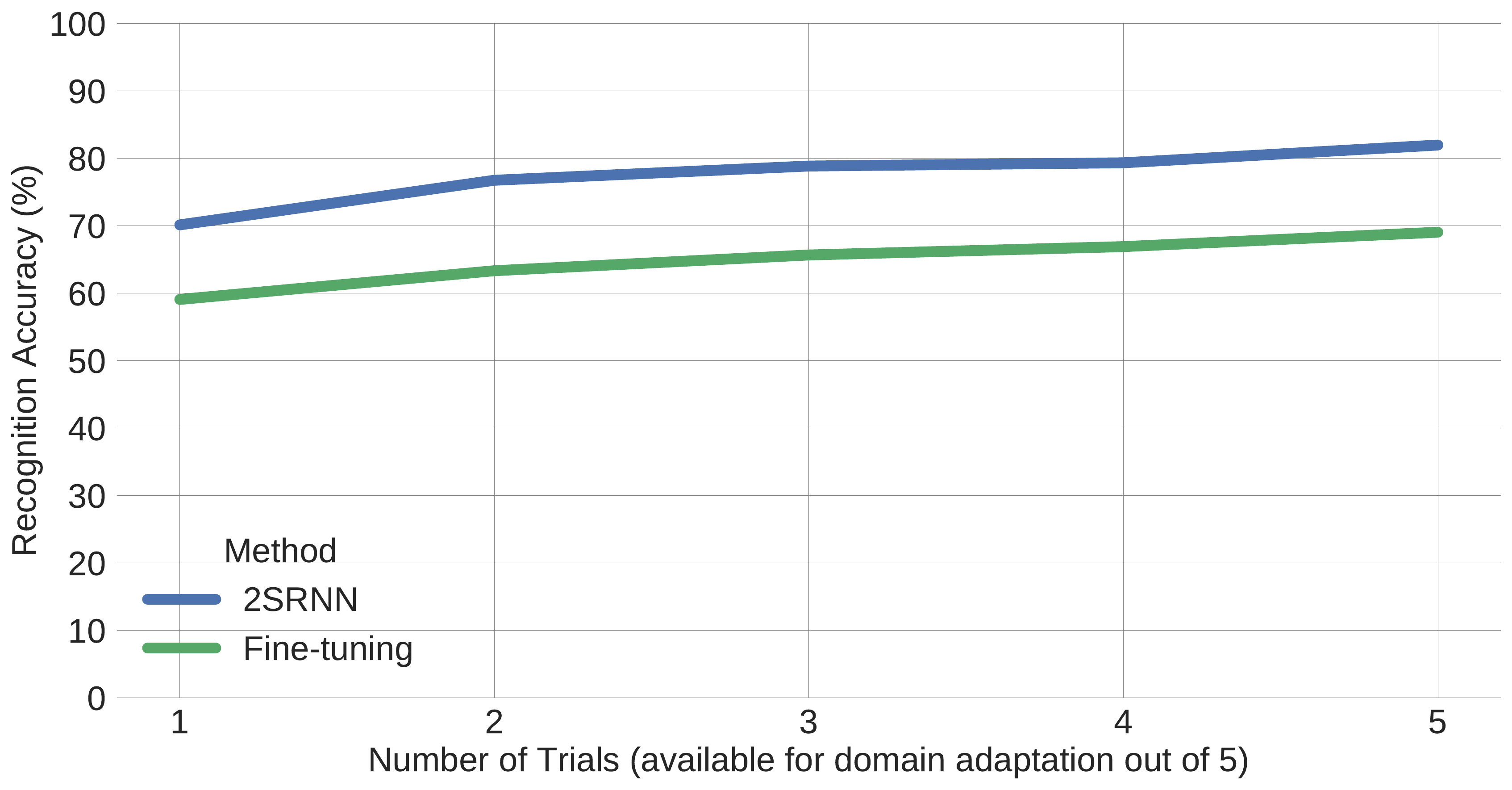}}
	\caption{Recognition accuracy comparison of supervised approaches}
	\label{figure:lineChart}
\end{figure}

It is natural to ask how much data is required to obtain a stable recognition accuracy and how our solutions relates to the common supervised fine-tuning method in deep learning.
Fig.~\ref{figure:lineChart} visualises a comparison of the inter-subject domain adaptation scenario (on the CapgMyo DB-b) based on our 2-stage RNN method and an adaptation based on supervised fine-tuning in one concrete scenario.
In this experiment we limited the available data to 20\%, 40\%, 60\%, 80\% and 100\%
of the total 5 trials used for domain adaptation (the remaining 5 trials are kept for validation). The mean classification accuracy is plotted as a function of the available target data for domain adaptation. Fig.~\ref{figure:lineChart} shows how the accuracy of the two method
increases with the amount of available target data and our 2SRNN remains persistently superior to the fine tuning method (by 20\%).
In each case we ran the domain adaptations for 5 epochs only since it is expected to get improvements quickly for better human-computer interactions.
On our server (with 2 Nvidia Titan V GPUs) these 5 training epochs took approximately 7.5 seconds for our 2SRNN and 27.7 seconds for supervised fine-tuning, respectively. Therefore a 20\% improvement in accuracy is complemented with a decrease in execution time by almost a factor of 4.

\section{Conclusions}

For real Human-Computer Interactions the sEMG-based gesture detection must overcome the inter-session and intersubject domain shifts. We proposed a 2-stage domain adaptation solution which has superior performance over the well-known supervised fine-tuning applied in deep learning and the state-of-the-art unsupervised adaptation methods. Empirical results show that the linear transformation of the input features is a good approximation for handling the domain shifts. It is fast and light weight and applicable (besides the RNN) to any machine learning approaches which are trainable with backpropagation. Combinations of this method with generative unsupervised models can be the next step of further improvements for usable HCI solutions.

The code is available at \url{https://github.com/ketyi/2SRNN}.

\end{document}